\pdfoutput=1

\documentclass[11pt]{article}

\usepackage{ACL2023}

\usepackage{times}
\usepackage{latexsym}

\usepackage[T1]{fontenc}

\usepackage[utf8]{inputenc}

\usepackage{microtype}

\usepackage{inconsolata}

\usepackage{graphicx}

\title{Deep Model Compression Also Helps Models Capture Ambiguity}

\author{Hancheol Park \hspace{6mm} Jong C. Park \\
        School of Computing \\ 
        Korea Advanced Institute of Science and Technology \\
        \texttt{\{hancheol.park,jongpark\}@kaist.ac.kr}}

\begin{document}
\maketitle
\begin{abstract}
Natural language understanding (NLU) tasks face a non-trivial amount of ambiguous samples where veracity of their labels is debatable among annotators. NLU models should thus account for such ambiguity, but they approximate the human opinion distributions quite poorly and tend to produce over-confident predictions. To address this problem, we must consider how to exactly capture the degree of relationship between each sample and its candidate classes. In this work, we propose a novel method with deep model compression and show how such relationship can be accounted for. We see that more reasonably represented relationships can be discovered in the lower layers and that validation accuracies are converging at these layers, which naturally leads to layer pruning. We also see that distilling the relationship knowledge from a lower layer helps models produce better distribution. Experimental results demonstrate that our method makes substantial improvement on quantifying ambiguity without gold distribution labels. As positive side-effects, our method is found to reduce the model size significantly and improve latency, both attractive aspects of NLU products.\footnote{Code is available at  \url{https://github.com/hancheolp/compression_for_capturing_ambiguity}.}
\end{abstract}

\section{Introduction}

Datasets constructed for natural language understanding (NLU) tasks, such as natural language inference (NLI) and text emotion analysis, contain a large amount of ambiguous samples \citep{nie-etal-2020-learn, uma2021}. As exemplified in Table~\ref{tab1}, each ambiguous sample is too debatable to be assigned a single gold label. Recent work has revealed that these disagreements among annotators are not annotation noise, which could have simply been resolved by aggregating more annotations, but rather a reproducible signal \citep{pavlick-kwiatkowski-2019-inherent, nie-etal-2020-learn}. This suggests that NLU models should predict not only majority labels, but also label distributions that respect such ambiguity.

\begin{table}
\centering
\begin{tabular}{l p{3.8cm}}
\hline
Premise & It’s summer time and two girls play with bubbles near a boat dock. \\
Hypothesis & It is warm outside. \\
Label distribution & Entailment: 0.52 \newline Neutral: 0.46 \newline Contradiction: 0.02 \\ 
\hline
News headline & Amateur rocket scientists reach for space. \\
Label distribution & Joy: 0.57 \newline Surprise: 0.43 \newline Anger / Disgust / Fear / Sadness: 0.00 \\ 
\hline
\end{tabular}
\caption{Ambiguous samples from datasets for NLI (ChaosSNLI \citep{nie-etal-2020-learn}) and emotion analysis (SemEval-2007 Task 14 dataset \cite{strapparava-mihalcea-2007-semeval})}
\label{tab1}
\end{table}

Since Transformer-based \citep{vaswani2017} pre-trained language models (PLMs) \citep{devlin-etal-2019-bert, liu2019} have become popular for NLU tasks, the accuracies of various NLU models have been substantially improved. Nevertheless, they are still not good at approximating the human opinion distributions \citep{pavlick-kwiatkowski-2019-inherent, nie-etal-2020-learn}, or label distributions drawn from a larger number of annotators, and their predictions tend to be over-confident \citep{zhang-etal-2021-learning-different}. If NLU products frequently produce over-confident predictions for ambiguous samples, it is not likely that they would be reliable for users who have different opinions.

As an attempt to address this problem, previous work \citep{zhang-etal-2021-learning-different, wang-etal-2022-capture} has demonstrated that label smoothing \citep{muller2019} helps make the prediction distributions close to human opinion distributions, simply addressing the issue of over-confidence. However, this does not explicitly address how to exactly capture the degree of relationship between each sample and its candidate classes (i.e., how to estimate $p(y=c|x)$ for each sample $x$). Some researchers \citep{zhang2018, meissner-etal-2021-embracing, wang-etal-2022-capture, zhou-etal-2022-distributed} have tried to use empirically-gold label distributions for directly learning the relationship, but these approaches require significant additional annotation costs.

In this paper, we propose a novel method that employs compression techniques for deep learning models, namely layer pruning \citep{sajjad2023} and knowledge distillation (KD) \citep{hinton2015}, and show how these compression techniques help models capture such a degree of relationship. We first observe that hidden states in lower layers more accurately encode the information about the sample-classes relationship, and that validation accuracies from internal classifiers inserted between adjacent layers are converging. This indicates that pruning a part of higher layers can make the models well represent the relationship information with their prediction distribution, while retaining the accuracy. We also observe that transferring the distribution knowledge that represents more accurate information about the relationship from a lower layer into the final classifier at the top of the pruned network can help the models produce better distribution.

Experimental results demonstrate that our method significantly outperforms existing ones that do not use additional distribution datasets. Without using such additional resources, our method also outperforms, or is comparable with, those that do use these resources over NLI benchmarks. Moreover, since our method uses compression techniques for deep learning models, this also reduces the model size significantly and improves latency as well. Both are attractive aspects of NLU products because they lead to consequent reduction in the cloud cost or to deployment on cheaper on-devices.

Deep model compression aims at eliminating redundant components of pre-trained deep learning models (via pruning or low-rank factorization \citep{liu2021}) to improve latency and reduce the model size. At the same time, maintaining the performance of the original model (via KD) is essential. While the goal of compression itself is not directly relevant to capturing ambiguity, we demonstrate that compression methods can also be used for accurately capturing ambiguity and suggest that such an approach presents another novel research direction for this task.

\section{Related Work}

\begin{figure*}
\centering
\includegraphics[width=1.0\linewidth]{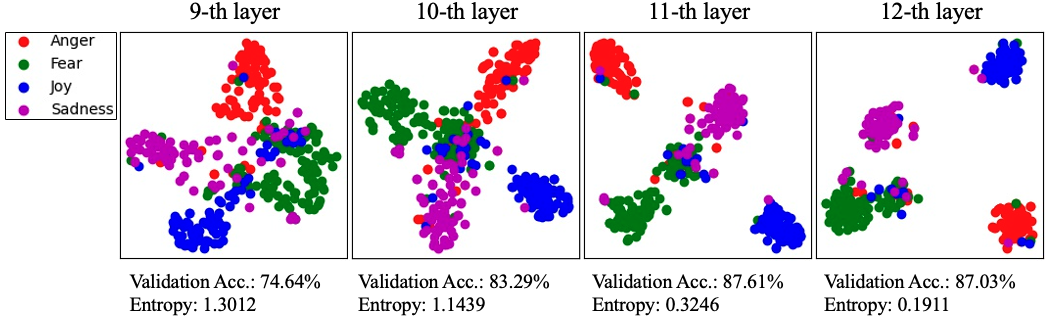}
\caption{Visualization of feature distributions from RoBERTa-base encoder layers using t-SNE}
\label{fig1}
\end{figure*}

Recent work has revealed that the state-of-the-art PLMs fine-tuned to predict single gold labels with cross-entropy loss function fail to properly estimate human opinion distributions \citep{pavlick-kwiatkowski-2019-inherent, nie-etal-2020-learn} and tend to produce over-confident predictions \citep{zhang-etal-2021-learning-different}. This issue of over-confidence is well-known in modern complex deep neural networks, because they can easily overfit one-hot labels of a training dataset. Moreover, this issue arises regardless of the correctness of predictions \citep{guo2017}.

In a situation where there exist a large number of ambiguous samples in an NLU dataset, it does not make sense to tolerate over-confident predictions. Naturally, in order to obtain better human opinion distributions, the use of label smoothing \citep{muller2019} has been proposed \citep{zhang-etal-2021-learning-different,wang-etal-2022-capture}. Label smoothing softens target training label distributions (i.e., one-hot labels) by shifting $\alpha$ probability mass from the target labels equally to all the labels. As a result, it prevents models from overfitting one-hot distribution. \citet{zhang-etal-2021-learning-different} and  \citet{wang-etal-2022-capture} have shown that label smoothing is effective at better estimating human opinion distributions. However, it makes all predictions less-confident, compared with using one-hot labels, not considering how to capture the degree of relationship between each sample and its candidate classes, which is an essential aspect to address ambiguity.

Monte Carlo dropout (MC dropout) \citep{gal2016} addresses the drawback of label smoothing. For a given sample, this method makes $k$ stochastic forward passes from a pre-trained neural network with dropout, where $k$ different prediction distributions are then averaged to form a final distribution for the sample. Since different forward passes could produce different plausible predictions for ambiguous samples, MC dropout also captures the aforementioned degree of relationship. Using MC dropout also improves the quality of output distributions \citep{zhou-etal-2022-distributed}, but this suffers from several drawbacks, such as its non-deterministic nature and higher latency for inference.

Directly learning from human opinion distributions has also been studied. \citet{zhang2018} and \citet{meissner-etal-2021-embracing} trained models with the empirically-gold label distributions to match predictions and human opinion distributions.  As post-editing, \citet{zhang-etal-2021-learning-different},  \citet{wang-etal-2022-capture}, and \citet{zhou-etal-2022-distributed} used temperature scaling \citep{guo2017}, with which output logits from a fine-tuned model are rescaled with hyperparameter $T$, and the softmax distribution becomes accordingly smoother and closer to target distributions.  $T$ is tuned on the distribution labels from a validation set by minimizing the KL-divergence between the predicted distributions and human opinion distributions. These additional resources significantly improve the ability to quantify ambiguity, but are accompanied with enormous annotation costs.

We propose to address all these limitations, considering how to exactly capture the degree of relationship between each sample and its candidate classes without the need for extra resources. In the next section, we explain how deep model compression can be made to account for the relationship without additional human opinion distribution information.

\section{Deep Model Compression for Capturing Ambiguity}
\subsection{Three Observations}
\label{sec3_1}

It is known that an average entropy value, measured from prediction distributions of an internal classifier inserted on top of each encoder layer, gradually becomes lower in the higher layers \citep{zhou2020}. However, it is not clear whether higher entropy values in the lower layers are attributed to the ability of those layers to encode ambiguous samples as high entropy distributions by assigning probabilities to all relevant classes. Therefore, we must look closely into how samples are encoded in each layer.

For this investigation, we use an emotion analysis dataset. This is because we can intuitively understand the relationship among emotion labels and such knowledge facilitates to interpret whether samples are well represented in accordance with our intuitions. We first fine-tuned RoBERTa-base \citep{liu2019} with an emotion analysis dataset, or “tweet emotion intensity dataset” \citep{mohammad-bravo-marquez-2017-emotion}. Each sample in this dataset was annotated via crowdsourcing with the intensity of its label (anger, fear, joy, or sadness). After fine-tuning, we froze all the parameters of the fine-tuned network and inserted a trainable internal classifier after every layer, which consists of the same layers with the original classifiers at the top layer. Finally, we trained the internal classifiers on the frozen network. In order to understand how the fine-tuned model encodes samples in each layer, we visualized the features of samples in the validation set, which are extracted from the hidden states for [CLS] tokens of layers (i.e., inputs of the internal classifiers), with t-SNE \citep{maaten2008}. In each layer, we also measured the validation accuracy using predictions from the internal classifiers and average entropy from predicted distributions on the same validation sets. The experimental results are shown in Figure ~\ref{fig1}.

We first observe that validation accuracy has already started to converge in lower layers (\textbf{observation (1)}). This result is identical to that of the previous work \citep{peters-etal-2019-tune}. Second, we observe that the feature representations from lower layers contain more accurate information about the degree of relationship between each sample and candidate classes (\textbf{observation (2)}). The relationship information visualized in Figure ~\ref{fig1} is considerably more intuitive and reasonable. In the 10-th layer, a sample from the ‘fear’ class is closely placed with samples in negative valance classes (i.e., ‘anger’ and ‘sadness’). In the next layer, a sample from the ‘fear’ class is distant from the ‘anger’ class, while close to the ‘sadness’ class that is highly correlated with the ‘fear’ class \citep{demszky-etal-2020-goemotions}. The internal classifier of the 11-th layer is usually likely to assign very low probabilities to the ‘anger’ class for samples from the ‘fear’ class. In the final layer, all classes are distantly located, to which the corresponding classifier is likely to make over-confident predictions. Intuitively, human annotators may recognize samples from ‘fear’ as ‘sadness’ or ‘anger’ with their subjective judgments, but such relationships disappear in the higher layers.

\begin{figure}
\centering
\includegraphics[width=0.9\linewidth]{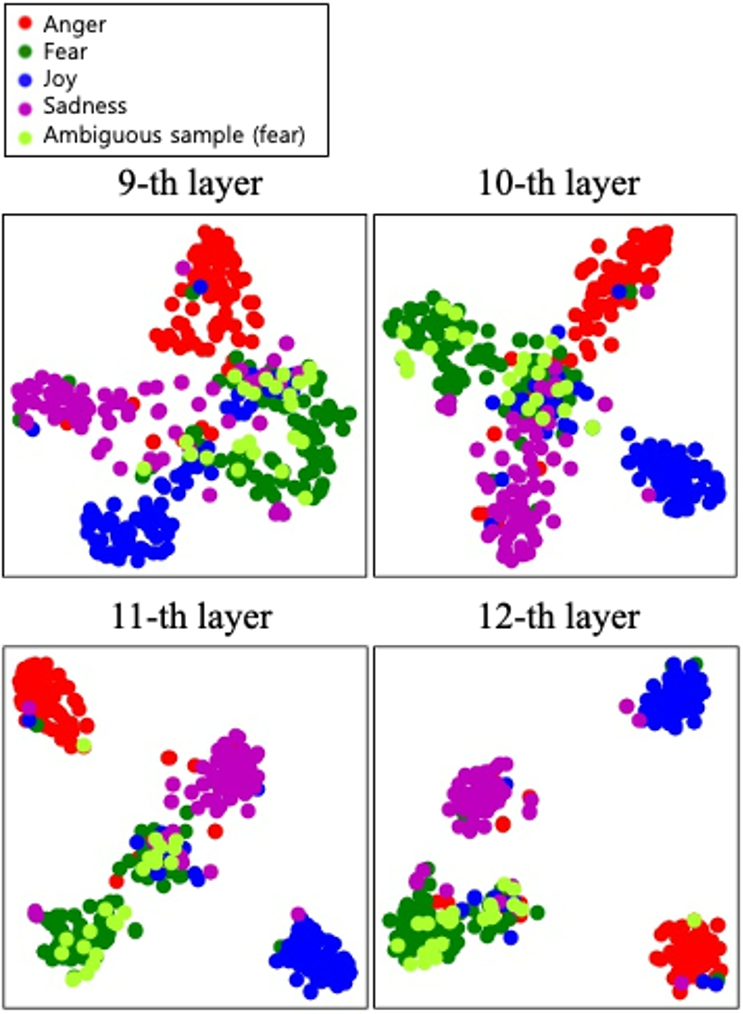}
\caption{Visualization of feature distributions for ambiguous samples that are labeled as ‘fear’}
\label{fig2}
\end{figure}

We further investigated how the model encodes ambiguous samples. We categorized samples depending on the emotional intensity scores (i.e., low: [0, 0.34), middle: [0.34, 0.67), high: [0.67, 1.0]) and assumed that samples that belong to a low intensity group are ambiguous. The underlying assumption is that an emotional tweet sample with extremely low intensity for its assigned class may also be relevant to other classes. As shown Figure ~\ref{fig2}, most of the ambiguous samples are closely placed with samples from their relevant classes, while non-ambiguous samples tend to be distantly located in the lower layers (i.e., 9-th and 10-th layers). However, in the higher layers, they seem not to be related with other classes anymore. Finally, we observe that after the most rapid drop of entropy values, each classifier starts to converge (\textbf{observation (3)}). These observations are made over BERT-base \citep{devlin-etal-2019-bert} as well (see Figure ~\ref{fig4} in Appendix~\ref{app1}).

\subsection{Layer Pruning}

From observations (1) and (2), we hypothesize that if we prune layers higher than the one where validation accuracy just starts to converge, we could obtain a model that better estimates the human opinion distributions, while retaining the performance. 

Given a fine-tuned PLM for NLU, all parameters of the model are frozen to maintain the encoded information about the relationship, and internal classifiers are then inserted between adjacent layers except at the top layer. This is the same procedure as used in our preliminary study in the previous section. Except for the final layer, which has already been fine-tuned, the internal classifiers are trained with the same configurations for training (e.g., the same training dataset and the same number of epochs) that are applied to the original fine-tuned PLM. Because we focus on the multi-class classification problem in NLU, the cross-entropy loss between predictions and gold labels (i.e., one-hot labels) is applied to all internal classifiers and the total loss function is $\sum_{i=1}^{n-1} L_i$ where $n$ is the total number of layers and $L_i$ is the cross-entropy loss function for the $i$-th internal classifier.

After training all internal classifiers, the validation accuracies from all classifiers are evaluated. Based on the evaluated accuracies, the target layer that will become the final layer after pruning should be selected. In this work, we simply select the lowest layer among those whose validation accuracy is higher than (the original validation accuracy – 1\%). We assumed that 1\% accuracy drop is tolerable in various NLU applications. In the case where a much higher accuracy is less important than well-quantified ambiguity, the threshold can be set higher than 1\% to prune more layers. After pruning layers above the target layer and removing the internal classifiers except for the last one at the top of the pruned network, we do not fine-tune the pruned model again. This is because we have already obtained the relationship information and training all parameters of the pruned model with one-hot labels turns the prediction from the model become over-confident again.

\subsection{Distilling the Relationship Information from a Lower Layer}
\label{sec3_3}

By layer pruning, our model could be made more accurate in terms of estimating human opinion distribution, but it should also be noted that when a model starts to converge, the entropy of prediction distribution has already decreased substantially (observation (3)) (i.e., prediction confidences would be significantly increased). This indicates that the pruned model may not be sufficient to produce a well-estimated human opinion distribution. In this case, pruning only yields improved distributions compared with models that are fine-tuned with one-hot labels, but does not let the model outperform previous methods. Therefore, if we further exploit the relationship information from much lower layers before the most rapid drop of entropy value, the pruned model could capture human opinion distributions more accurately. In order to transfer such knowledge to the classifier layer of the pruned network, we propose a variant of knowledge distillation (KD) \citep{hinton2015}.

Originally, KD is a training technique to recover the accuracy of a compressed or smaller model (i.e., student model) using the knowledge (e.g., output distributions or feature representations) from the original or a larger model (i.e., teacher model), which is more accurate than the student model. The goal of KD is to match the prediction distribution (or feature representations) from the student with that from the teacher. In our case, the knowledge should provide more accurate information about the relationship between each sample and candidate classes. Therefore, in this work, we transfer the prediction distributions from a lower layer into the final classifier of the pruned layer, which is an approach different from the conventional one.

In this work, we set the layer just before the most rapid drop of entropy on the pruned network as the source layer that transfers the distribution information, because the distribution information from much lower layers can degrade the accuracy. The entropy can be measured with a validation set before removing internal classifiers in the previous pruning step. In order not to change the distribution information during KD, we froze the parameters by the source layer and updated the parameters above the source layer to adjust the prediction distribution of the last layer. The loss function $L_{kd}$ for our KD approach is computed as follows:
\begin{equation}
L_{kd} = \lambda L_{ce}(\bar{y}_t, y) + (1 - \lambda) L_{ce}(\bar{y}_t, \bar{y}_s)
\end{equation}
where $y$ is one-hot labels, $\bar{y}_t$ is prediction distributions from the target layer, $\bar{y}_s$ is prediction distributions from the source layer, and $L_{ce}$ is the cross-entropy loss function. The first term on the right side is used to avoid the accuracy drop from incorrect majority label information that exists in $\bar{y}_s$. The second term is a distillation loss, which makes output distributions close to the distributions from the source layer. $\lambda$ is a hyperparameter that determines the quantity of the transferred knowledge from the source layer. A smaller value of $\lambda$  could lead to a broader incorporation of relationship information, but it may result in a less accurate model. Therefore, it is important to find an optimal $\lambda$ for a model that can estimate human opinion distributions accurately, while retaining the performance. However, it is challenging to tune $\lambda$ since a validation set that contains gold label distributions is not available in our setting (i.e., using only single gold labels). If we have such a dataset, we could easily find $\lambda$ by investigating the distance between predictive distributions and gold label distributions. 

To address the issue of hyperparameter tuning, we propose a sub-optimal solution as follows. First, $\lambda$ should be larger than 0.5 to give more weight to correct learning signals over noisy ones when $\bar{y}_s$ represents incorrect labels. Second, we select $\lambda$ in such a way that the validation accuracy of the model is higher than the original validation accuracy minus 1\%. Finally, we determine $\lambda$ with which the average prediction probability for ground truth labels in a validation set is maximum when predictions are incorrect. This choice is made because assigning high probability values to the ground truth labels, even when the predictions are incorrect, helps to minimize the discrepancy between the model's outputs and the true human opinion distributions (e.g., when the ground truth distribution is [\underline{0.6}, 0.35, 0.05], the prediction [\underline{0.4}, 0.55, 0.05] is closer to the true distribution than predicted [\underline{0.2}, 0.75, 0.05]). Moreover, since maximizing probabilities to the ground truth labels naturally leads to decreasing the probabilities to incorrect labels, we can avoid the risk that strongly favors incorrect predictions. In our experiments, we tuned $\lambda$ with the candidate values \{0.6, 0.7, 0.8, 0.9\}. We also trained models with the same configurations for training that are applied to the original fine-tuned PLM.

\section{Experiments}

In this section, we investigate how exactly our method of using compression techniques can capture the ambiguity of each sample without empirically-gold label distributions. We re-ran all experiments three times with different random seeds to identify variance. The standard deviation value of accuracy is smaller than 0.0155 on all methods and datasets and of Jenson-Shannon Distance (JSD) \citep{endres2003} is smaller than 0.0081, both of which are negligible. 

\subsection{Metrics}

In order to examine to what extent models are capable of capturing ambiguity, we use JSD as a primary metric, which measures the distance between the softmax outputs of the models and the gold human label distributions. Since this metric is symmetric and bounded with the range [0, 1], it has been  popularly used in the previous work \citep{nie-etal-2020-learn, zhang-etal-2021-learning-different, meissner-etal-2021-embracing, wang-etal-2022-capture, zhou-etal-2022-distributed}. We also use KL divergence to measure the distance as a complementary metric due to its limitation (i.e., non-symmetry).

\subsection{Baseline Methods}

We first compare our method with baselines that use the same single gold labels for training, such as the standard training method (STD) (i.e., training with one-hot labels and cross-entropy loss function), MC dropout (MC) \citep{zhou-etal-2022-distributed}, and label smoothing (LS) \citep{zhang-etal-2021-learning-different, wang-etal-2022-capture}. For MC dropout, we set the dropout probability to 0.1, which is the value for pre-training the language model used in our experiments and $k$ to 10. $\alpha$ of the label smoothing is set to 0.1 because it tends to be set as 0.1 over many datasets \citep{muller2019}. 

We also compare our method with baselines that use additional human opinion distribution datasets, such as temperature scaling (TS) \citep{zhou-etal-2022-distributed, wang-etal-2022-capture} and label distribution learning (LDL) (i.e., training with human opinion distributions and cross-entropy loss function) \citep{zhang2018, meissner-etal-2021-embracing}.  We also report the results from the chance baseline. For the chance baseline, JSD and KL divergence between uniform distributions and human opinion distributions are calculated. Accuracy is the proportion of the samples to the majority label in each test set.

\begin{table*}
\centering
\begin{tabular}{l c c c c c c c c c}
\hline
& \multicolumn{3}{c}{\textbf{ChaosSNLI}} & \multicolumn{3}{c}{\textbf{ChaosMNLI}} & \multicolumn{3}{c}{\textbf{Emotion}}\\
& \textbf{JSD$\downarrow$} & \textbf{KL$\downarrow$} & \textbf{Acc.$\uparrow$} 
& \textbf{JSD$\downarrow$} & \textbf{KL$\downarrow$} & \textbf{Acc.$\uparrow$}
& \textbf{JSD$\downarrow$} & \textbf{KL$\downarrow$} & \textbf{Acc.$\uparrow$} \\
\hline
\textbf{Chance}   & 0.3829 & 0.5456 & 0.5370 & 0.3022 & 0.3558 & 0.4634 & 0.4728 & 0.8588 & 0.3211 \\
\textbf{STD}      & 0.3299 & 1.3872 & 0.6935 & 0.4219 & 2.3982 & \textbf{0.5722} & 0.4203 & 1.2858 & 0.5528 \\
\textbf{MC}       & 0.2984 & 0.9287 & 0.6849 & 0.3718 & 1.6320 & 0.5710 & 0.4044 & 1.0381 & 0.5203 \\
\textbf{LS}       & 0.2723 & 0.5724 & 0.7173 & 0.3540 & 0.8574 & 0.5591 & 0.4057 & 0.9825 & \textbf{0.5610} \\
\hline            
\textbf{TS}       & 0.2626 & 0.5099 & 0.6935 & 0.3095 & 0.6491 & \textbf{0.5722} & 0.3859 & 0.7708 & 0.5528 \\
\textbf{LDL}      & \textbf{0.2185} & 0.3811 & \textbf{0.7186} & 0.2991 & 0.7032 & 0.5716 & \textbf{0.3338} & \textbf{0.5198} & \textbf{0.5610} \\
\hline
\textbf{Ours}     & 0.2635 & \textbf{0.3642} & 0.7127 & \textbf{0.2799} & \textbf{0.4707} & 0.5691 & 0.3935 & 0.8703 & 0.5447 \\
\hline
\end{tabular}
\caption{\label{tab2}
Evaluation results for methods on JSD, KL, and Acc. $\downarrow$ indicates that a smaller value is better. $\uparrow$ indicates that a larger value is better. The best values among methods are highlighted in bold.
}
\end{table*}

\begin{table}
\centering
\begin{tabular}{l p{1.3cm} p{1.3cm} p{1.3cm}} 
\hline
& {\textbf{Chaos \newline SNLI}} & {\textbf{Chaos \newline MNLI}} & {\textbf{Emotion}} \\
\hline
\textbf{STD}   & 0.3299 & 0.4219 & 0.4203 \\
\hline
\textbf{+ Pruning} & 0.3197 & 0.4091 & 0.4069 \\
\textbf{+ KD}      & 0.2672	& 0.2881 & 0.3981 \\
\textbf{+ All}     & 0.2635	& 0.2799 & 0.3935 \\
\hline            
\end{tabular}
\caption{\label{tab3}The results of ablation study (metric: JSD)}
\end{table}

\subsection{Datasets}

In this work, we use datasets for NLI and text emotion analysis. As test sets of the NLI task, we used ChaosMNLI (1,599 MNLI-matched development set \citep{williams-etal-2018-broad}) and ChaosSNLI datasets (1,514 SNLI development set \citep{bowman-etal-2015-large}) \citep{nie-etal-2020-learn}. In these datasets, each sample was labeled by 100 annotators and these annotations were normalized to represent human opinion distributions. As training and validation sets, we used AmbiSM datasets \citep{meissner-etal-2021-embracing}. AmbiSM provides empirically-gold label distributions collected by crowd-sourcing annotation. AmbiSM consists of SNLI development/test set and MNLI-matched/mismatched development set, in which none of the samples overlaps with those in ChaosNLI. When models are evaluated with ChaosMNLI, we used randomly selected 1,805 MNLI-matched development samples in AmbiSM, as validation set and the rest of AmbiSM were used as training set (34,395 samples). For ChaosSNLI, we used 1,815 SNLI development samples as validation set and the rest of AmbiSM were used as training set (34,385 samples).

For text emotion analysis, we used SemEval-2007 Task 14 Affective Text dataset \citep{strapparava-mihalcea-2007-semeval}. We used 800 samples for training, 200 for validation, and 246 for evaluation (4 “neutral” labels were excluded from evaluation.). In this dataset, 6 emotion intensities (i.e., anger, disgust, fear, joy, sadness, and surprise) are labeled by annotators and each intensity value is normalized to get label distributions using the same procedure as in the previous work \citep{zhang2018}.

\subsection{Implementation Details}

Our proposed method and baselines are applied to RoBERTa-base \citep{liu2019}. The implementation of RoBERTa-base was based on Huggingface Transformers\footnote{\url{https://github.com/huggingface/transformers}}. All methods used the same hyperparameters for training. Batch size was 32, and learning rate was 5e-5 with a linear decay. We fine-tuned over 5, 6, and 7 epochs for ChaosSNLI, ChaosMNLI, and the emotion dataset, respectively, based on the validation accuracy. We used AdamW optimizer \citep{Loshchilov2019} for parameter update. Weight decay was set to 0.1.

\subsection{Results}

We describe the experimental results that are measured on the test sets in Table ~\ref{tab2}. As researchers demonstrated in the previous work \citep{pavlick-kwiatkowski-2019-inherent, nie-etal-2020-learn}, the standard method poorly estimates human opinion distributions and does not always outperform the chance baseline. On the other hand, our method significantly outperforms all baseline methods that are trained with single gold labels (STD, MC, and LS). Moreover, for NLI tasks, our proposed method outperforms or is comparable with the baseline method that uses additional human opinion distribution datasets. However, for the emotion dataset, our method does not outperform the methods that use the additional resource.

These experimental results suggest that the relationship information encoded in the lower layers is also a useful source for estimating human opinion distributions. Moreover, such relationship information could be more accurate than the relationship information obtained from different forward passes from MC dropout.

\begin{table*}
\centering
\begin{tabular}{l c c c c c c}
\hline
& \multicolumn{2}{c}{\textbf{ChaosSNLI}} & \multicolumn{2}{c}{\textbf{ChaosMNLI}} & \multicolumn{2}{c}{\textbf{Emotion}}\\
& \textbf{Diff.$\downarrow$} & \textbf{JSD$\downarrow$}  
& \textbf{Diff.$\downarrow$} & \textbf{JSD$\downarrow$} 
& \textbf{Diff.$\downarrow$} & \textbf{JSD$\downarrow$} \\
\hline
\textbf{STD}      & 0.6092 & 0.3299 & 0.5749 & 0.4219 & 0.3987 & 0.4203 \\
\textbf{MC}       & 0.5476 & 0.2984 & 0.5187 & 0.3718 & 0.3711 & 0.4044 \\
\textbf{LS}       & 0.5753 & 0.2723 & 0.5469 & 0.3540 & 0.3850 & 0.4057 \\
\textbf{TS}       & 0.5342 & 0.2626 & 0.4957 & 0.3095 & 0.3663 & 0.3859 \\
\textbf{LDL}      & 0.4819 & 0.2185 & 0.4435 & 0.2991 & 0.2866 & 0.3338 \\
\hline
\textbf{Pruning}  & 0.5997 & 0.3197 & 0.5614 & 0.4091 & 0.3791 & 0.4069 \\
\textbf{+KD}      & 0.5265 & 0.2635 & 0.4686 & 0.2799 & 0.3610 & 0.3935 \\
\hline
\end{tabular}
\caption{\label{tab4}
The relationship between JSD and the average difference between the ground truth probabilities and predicted probabilities for the ground truth labels (Diff.) when predictions are incorrect.
}
\end{table*}

\begin{table*}
\centering
\begin{tabular}{l c c c c c c c c c}
\hline
& \multicolumn{3}{c}{\textbf{ChaosSNLI}} & \multicolumn{3}{c}{\textbf{ChaosMNLI}} & \multicolumn{3}{c}{\textbf{Emotion}}\\
& \textbf{JSD$\downarrow$} & \textbf{KL$\downarrow$} & \textbf{Acc.$\uparrow$} 
& \textbf{JSD$\downarrow$} & \textbf{KL$\downarrow$} & \textbf{Acc.$\uparrow$}
& \textbf{JSD$\downarrow$} & \textbf{KL$\downarrow$} & \textbf{Acc.$\uparrow$} \\
\hline
\textbf{LS}      & 0.2723 & 0.5724 & 0.7173 & 0.3540 & 0.8574 & 0.5591 & 0.4057 & 0.9825 & 0.5610 \\
\textbf{LS+Ours} & 0.2441 & 0.3413 & 0.7200 & 0.2603 & 0.3786 & 0.5653 & 0.3800 & 0.7137 & 0.5569 \\
\hline
\end{tabular}
\caption{\label{tab5}
The degree of improvement when our method is applied to the fine-tuned models with label smoothing
}
\end{table*}

\section{Discussion}

\begin{figure}
\centering
\includegraphics[width=1.0\linewidth]{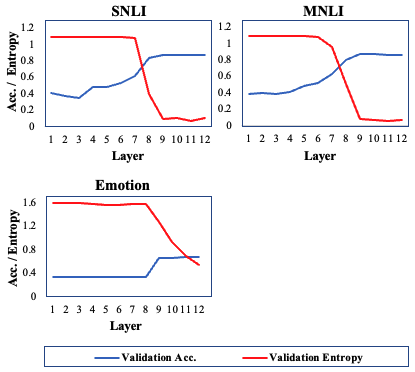}
\caption{Validation accuracy and entropy in all layers.}
\label{fig3}
\end{figure}

\textbf{Which compression method is more effective?} As described in Table ~\ref{tab3}, KD is the most effective technique to capture the ambiguity. As we argued in Section ~\ref{sec3_3}, even though applying only the layer pruning technique can yield better distributions than the STD, it is not sufficient to exactly capture the relationship information. Nevertheless, this technique is still helpful to improve the ability to capture the ambiguity when KD is used together. Therefore, in a situation where highly reliable and faster models are required, pruning can be a good option.
\newline \newline
\textbf{Can our observations be reproduced over different datasets?} The design of our proposed methods is based on our three observations in Section ~\ref{sec3_1}. If these do not manifest in other datasets, our method may not work in general. Therefore, we conducted the same procedure described in Section ~\ref{sec3_1} on the datasets used in our experiments. As described in Figure ~\ref{fig3}, in these datasets, we observed that the validation accuracy is starting to converge in the lower layers (observation (1)). We also found that feature representations from a lower layer contain richer information about the degree of relationship (see Figure ~\ref{fig5} in Appendix ~\ref{app1}). We also observed again that after the most rapid drop of entropy value occurs, models started to converge as described in Figure ~\ref{fig3}. These suggest that our method can be applied to various other NLU datasets as well. 
\newline \newline
\textbf{Is maximizing the probabilities for ground truth labels when predictions are incorrect a valid solution for tuning $\lambda$ of our KD loss?} In order to validate the tuning approach, we measure the average difference between the ground truth probabilities and predicted probabilities for the ground truth labels as described in Table ~\ref{tab4}. We found that KD with our tuning approach significantly reduces the differences by maximizing the probabilities for the ground truth labels and the reduced differences tend to decrease the values of JSD, which suggests that our proposed tuning approach is valid.
\newline \newline
\textbf{Can models trained with label smoothing be improved with our proposed method?} Since our method is applied to a fine-tuned model, we looked into whether the proposed method can further improve the estimation ability for human opinion distributions on models trained with label smoothing. In this case, we used smoothed labels instead of one-hot labels $y$ during knowledge distillation. As shown in Table ~\ref{tab5}, our method can significantly improve the ability of capturing ambiguity in the models that have already been calibrated with label smoothing.
\newline \newline
\textbf{What are additional benefits of our method?} In our experiments, 1 layer is pruned for the emotion analysis model and 3 layers are pruned for the NLI models. These result in significant reduction in the number of model parameters (from 125M (RoBERTa-base) to 117M and to 103M, respectively). We also measured the average latency per 300 token input on a low-end mobile device (i.e., Samsung Galaxy Tab S6 Lite). The pruned network is also found to significantly reduce the latency on the mobile device (from 2.42 sec. to 2.22 sec. and to 1.86 sec., respectively).

\section{Conclusion}

In this work, we proposed a novel method for capturing ambiguity with deep model compression techniques, namely layer pruning and knowledge distillation. Experimental results demonstrate that our method substantially improves the ability of quantifying ambiguity and provides efficient compressed models for NLU products.

As future work, we would further investigate the availability of different compression methods such as pruning self-attention heads and FNN because redundant components in modern complex deep learning may lead to over-confidence \citep{guo2017}. In another direction, we may also address limitations that are revealed in our work, such as multiple training procedures or hyperparameter tuning for each method (e.g., how much we allow accuracy drop during layer pruning).

\section*{Limitations}

Although our method well estimates the ambiguity without additional resources as well as boosting model latency significantly, there are a few limitations. First, our method requires additional training procedures, such as training the internal classifiers and KD. For this, we may fine-tune the original model and internal classifiers simultaneously. Another limitation is in setting the hyperparameters. We allow the drop of accuracy by 1\% to determine the target layer for layer pruning and the value of $\lambda$ for KD, but this could be subjective and differ depending on the researchers’ experience. Finally, we validated our method with a limited number of benchmarks since most of datasets have been released with only  aggregated gold labels \citep{uma2021}.

\section*{Ethics Statement}
We used well-known datasets that have no ethical issues (S/MNLI and SemEval-2007 Task 14 dataset). However, some samples may contain contents unsuitable for certain individuals. In particular, the SemEval-2007 Task 14 dataset  provides news headlines that evoke readers’ negative emotional reaction. 

It should also be noted that our method cannot still produce completely reliable distributions. This means that our method may suffer from false facts or biases. There is thus a possibility that one can misuse our model to support their false facts with the results from our model, though problems of this kind are not unique to our model.

\section*{Acknowledgements}
This work was supported by Institute for Information and communications Technology Promotion (IITP) grant funded by the Korea government (No. 2018-0-00582, Prediction and augmentation of the credibility distribution via linguistic analysis and automated evidence document collection).

\bibliography{anthology,custom}
\bibliographystyle{acl_natbib}

\newpage
\appendix
\section{Reproducibility of Our Observations over Different Models and Datasets}
\label{app1}

\begin{figure}[htb!]
\centering
\includegraphics[width=0.8\linewidth]{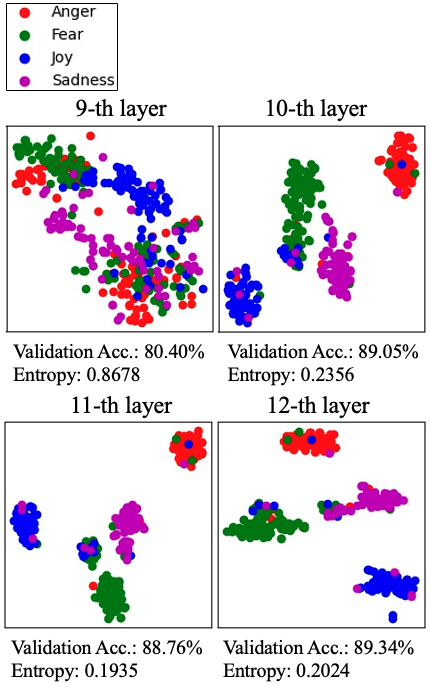}
\caption{Visualization of feature distributions on tweet emotion intensity dataset samples in BERT-base layers}
\label{fig4}
\end{figure}

\begin{figure}[htb!]
\centering
\includegraphics[width=0.8\linewidth]{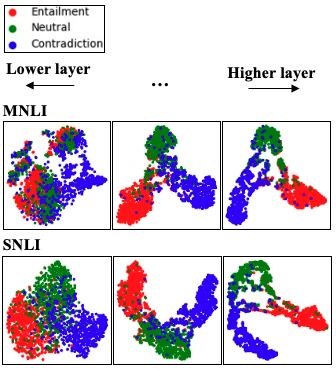}
\caption{Visualization of feature distributions on the validation sets of MNLI (top) and SNLI (bottom)}
\label{fig5}
\end{figure}

\end{document}